%% file: main.tex
\documentclass{article}
\usepackage{ijcai21}

\usepackage{times}
\usepackage{soul}
\usepackage{url}
\usepackage[hidelinks]{hyperref}
\usepackage[utf8]{inputenc}
\usepackage[small]{caption}
\usepackage{graphicx}
\usepackage{amsmath}
\usepackage{amsthm}
\usepackage{tikz}
\usepackage{pgfplots}
\usepackage{booktabs}
\usepackage{algorithm}
\usepackage{algorithmic}
\input{math_commands.tex}

\usetikzlibrary{arrows,arrows.meta}

\def\parspace{4pt}

\title{Neural Temporal Point Processes: A Review}

\author{
Oleksandr Shchur$^{1,2}$
\and
Ali Caner T\"{u}rkmen$^2$\and
Tim Januschowski$^{2}$\And
Stephan G\"{u}nnemann$^1$
\affiliations
$^1$Technical University of Munich, Germany\\
$^2$Amazon Research
\emails
\{shchur,guennemann\}@in.tum.de,
\{atturkm,tjnsch\}@amazon.com,
}

\begin{document}

\maketitle

\begin{abstract}
Temporal point processes (TPP) are probabilistic generative models for continuous-time event sequences.
Neural TPPs combine the fundamental ideas from point process literature with deep learning approaches, thus enabling construction of flexible and efficient models.
The topic of neural TPPs has attracted significant attention in recent years, leading to the development of numerous new architectures and applications for this class of models.
In this review paper we aim to consolidate the existing body of knowledge on neural TPPs.
Specifically, we focus on important design choices and general principles for defining neural TPP models.
Next, we provide an overview of application areas commonly considered in the literature. 
We conclude this survey with the list of open challenges and important directions for future work in the field of neural TPPs.

\end{abstract}

\begin{figure*}
    \centering
    \begin{tikzpicture}
    \node (image) at (0,0) 
    {\includegraphics[width=0.9\textwidth]{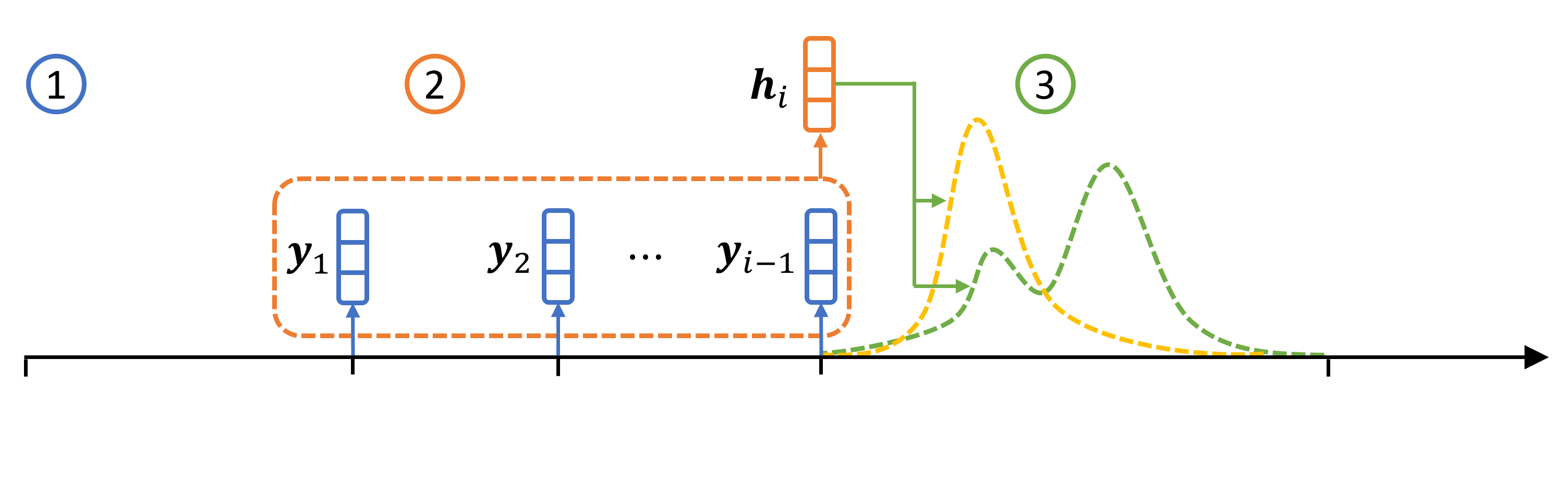}};
    \node[align=center] (c) at (-5.8, 1.5) {\small Represent events\\\small $(t_j, m_j)$ as feature\\\small vectors $\vy_j$};
    \node[align=center] (c) at (-2.08, 1.5) {\small Encode history\\\small $(\vy_1, \dots, \vy_{i-1})$\\\small as a vector $\vh_i$};
    \node[align=center] (c) at (5.2, 1.73) {\small Obtain conditional distribution\\\small $P_i(t_i, m_i | \history_{t_i}) = P(t_i, m_i | \vh_i)$};
    \node[align=center] (c) at (-7.72, -1.6) {\small $0$};
    \node[align=center] (c) at (-4.4, -1.6) {\small $(t_1, m_1)$};
    \node[align=center] (c) at (-2.3, -1.6) {\small $(t_2, m_2)$};
    \node[align=center] (c) at (0.40, -1.6) {\small $(t_{i-1}, m_{i-1})$};
    \node[align=center] (c) at (-1.05, -1.6) {\small $\cdots$};
    \node[align=center] (c) at (5.6, -1.6) {\small $T$};
    \node[align=center] (c) at (7.5, -1.6) {\small time};
    \end{tikzpicture}
    \vspace{-7mm}
    \caption{Schematic representation of an autoregressive neural TPP model.}
    \label{fig:ar-tpp}
\end{figure*}
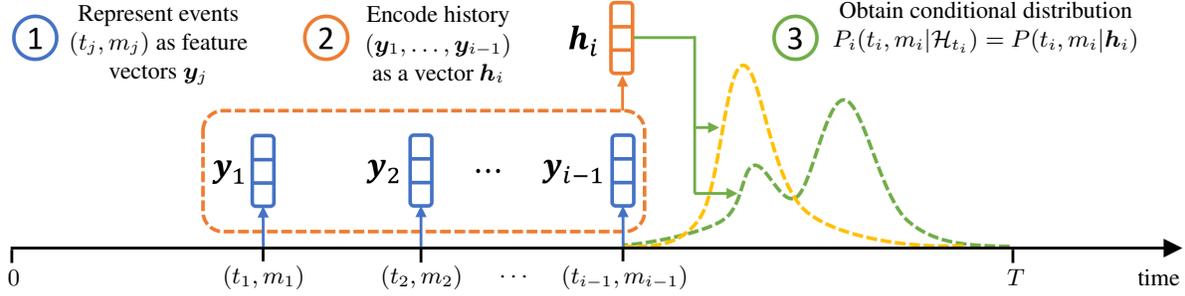

\section{Introduction}

Many applications in science and industry are concerned with collections of
events with \emph{timestamps}.
Earthquake occurrences in seismology, neural spike trains in
neuroscience, trades and orders in a financial market, and user activity logs
on the web, can all be represented as sequences of discrete (instantaneous) events observed in continuous time.

Temporal point processes (TPP) are probabilistic models for such event data
\cite{daley2007introduction}.
More specifically, TPPs are generative models of variable-length point
sequences observed on the real half-line---here interpreted
as arrival times of events.
TPPs are built on rich theoretical foundations, with early work dating back
to the beginning of the 20th century, where they were used to model the arrival
of insurance claims and telephone traffic \cite{brockmeyer1948life,cramer1969historical}.
The field underwent rapid development in the second half of the century, and TPPs were
applied to a wide array of domains including seismology, neuroscience, and
finance.

Nevertheless, TPPs entered the mainstream of machine learning research only very recently.
One of the exciting ideas developed at the intersection of the fields of point processes and machine learning were \emph{neural} TPPs \cite{du2016recurrent,mei2017neural}.
Classical (non-neural) TPPs can only capture relatively simple patterns in event occurrences, such as self-excitation \cite{hawkes1971point}.
In contrast, neural TPPs are able to learn complex dependencies, and are often even computationally more efficient than their classical counterparts.
As such, the literature on neural TPPs has witnessed rapid growth since their introduction.

\vspace{\parspace}
\noindent\textbf{Scope and structure of the paper.}
The goal of this survey is to provide an overview of neural TPPs, with focus on models (Sections~\ref{sec:autoregressive}--\ref{sec:learning}) and their applications (\secref{sec:applications}). 
Due to limited space, we do not attempt to describe every existing approach in full detail, but rather focus on general principles and building blocks for constructing neural TPP models.

We also discuss the main challenges that the field currently faces and outline some future research directions (\secref{sec:discussion}).
For other reviews of TPPs for machine learning, we refer the reader to the tutorial by \cite{rodriguez2018learning};
and two recent surveys by \cite{yan2019recent}, who also covers non-neural approaches,
and \cite{enguehard2020neural} who experimentally compare neural TPP architectures in applications to healthcare data. 
Our work provides a more detailed overview of neural TPP architectures and their applications compared to the above papers.

\section{Background and Notation}
\label{sec:notation}
A TPP \cite{daley2007introduction} is a probability distribution over variable-length sequences in some time interval $[0, T]$.
A realization of a \emph{marked} TPP can be represented as an event sequence $X = \{(t_1, m_1), \dots, (t_N, m_N)\}$, where $N$, the number of events, is itself a random variable.
Here, $0 < t_1 < \dots < t_N \le T$ are the arrival times of events and $m_i \in \gM$ are the marks.
Categorical marks (\ie, $\gM = \{1, \dots, K\}$) are most commonly considered in practice, but other choices, such as $\gM = \R^D$, are also possible.
Sometimes, it is convenient to instead work with the inter-event times $\tau_i = t_i - t_{i-1}$, where $t_0 = 0$ and $t_{N+1} = T$.
For a given $X$, we denote the history of past events at time $t$ as $\history_t = \{(t_j, m_j): t_j < t\}$.

A distribution of a TPP with $K$ categorical marks can be characterized by $K$ 
conditional intensity functions $\cintensity_k(t)$ (one for each mark $k$) that are defined as
\begin{align}
    \label{eq:intensity}
    \cintensity_k(t) = \lim_{\Delta t \downarrow 0} \frac{\Pr(\text{event of type $k$ in } [t, t+ \Delta t)| \history_t)}{\Delta t},
\end{align}
where the $*$ symbol is used as a shorthand for conditioning on the history $\history_t$.
Note that the above definition also applies to unmarked TPPs if we set $K=1$.
While the conditional intensity is often mentioned in the literature, it is not the only way to characterize a TPP---we will consider some alternatives in the next section.

\section{Autoregressive Neural TPPs}
\label{sec:autoregressive}
Neural TPPs can be defined as autoregressive models, as done in the seminal work by \cite{du2016recurrent}.
Such autoregressive TPPs operate by sequentially predicting the time and mark of the next event $(t_i, m_i)$.
Usually, we can decompose this procedure into 3 steps (see \figref{fig:ar-tpp}):
\begin{enumerate}
    \item Represent each event $(t_j, m_j)$ as a feature vector $\feat_j$. 
    \item Encode the history $\history_{t_i}$ (represented by a sequence of feature vectors $(\feat_1, \dots, \feat_{i-1})$) into a fixed-dimensional history embedding $\hemb_i$. 
    \item Use the history embedding $\hemb_i$ to parametrize the conditional distribution over the next event $\P_i(t_i, m_i | \history_{t_i})$.
\end{enumerate}
We will now discuss each of these steps in more detail.

\subsection{Representing Events as Feature Vectors}
\label{sec:ar-feat}
First, we need to represent each event $(t_j, m_j)$ as a feature vector $\feat_j$ that can then be fed into an encoder neural network (\secref{sec:ar-encode}).
We consider the features $\timefeat_j$ based on the arrival time $t_j$ (or inter-event time $\tau_j$) and $\markfeat_j$ based on the mark $m_j$.
The vector $\feat_j$ is obtained by combining $\timefeat_j$ and $\markfeat_j$, \eg, via concatenation.

\vspace{\parspace}
\noindent\textbf{Time features} $\timefeat_j$.
Earlier works used the inter-event time $\tau_j$ or its logarithm $\log \tau_j$ as the time-related feature \cite{du2016recurrent,omi2019fully}.
Recently, \cite{zuo2020transformer} and \cite{zhang2020self} proposed to instead obtain features from $t_j$ using trigonometric functions, which is based on positional encodings used in transformer language models \cite{vaswani2017attention}.

\vspace{\parspace}
\noindent
\textbf{Mark features} $\feat^{\textrm{mark}}_j$.
Categorical marks are usually encoded with an embedding layer \cite{du2016recurrent}.
Real-valued marks can be directly used as $\markfeat_j$.
This ability to naturally handle different mark types is one of the attractive properties of neural TPPs (compared to classical TPP models).

\subsection{Encoding the History into a Vector}
\label{sec:ar-encode}
The core idea of autoregressive neural TPP models is that event history $\history_{t_i}$ (a variable-sized set) can be represented as a \emph{fixed-dimensional} vector $\hemb_i$ \cite{du2016recurrent}.
We review the two main families of approaches for encoding the past events $(\feat_1, \dots, \feat_{i-1})$ into a history embedding $\hemb_i$ next.

\vspace{\parspace}
\noindent \textbf{Recurrent encoders} 
start with an initial hidden state $\hemb_1$.
Then, after each event $(t_i, m_i)$ they update the hidden state as $\hemb_{i+1} = \operatorname{Update}(\hemb_{i}, \feat_{i})$.
The hidden states $\hemb_i$ are then used as the history embedding.
The $\operatorname{Update}$ function is usually implemented based on the RNN, GRU or LSTM update equations \cite{du2016recurrent,xiao2017modeling}.

The main advantage of recurrent models is that they allow us to compute the history embedding $\hemb_i$ for all $N$ events in the sequence in $O(N)$ time.
This compares favorably even to classical non-neural TPPs, such as the Hawkes process, where the likelihood computation in general scales as $O(N^2)$.
One downside of recurrent models is their inherently sequential nature.
Because of this, such models are usually trained via truncated backpropagation through time, which only provides an approximation to the true gradients \cite{sutskever2013training}.

\vspace{\parspace}
\noindent \textbf{Set aggregation encoders} 
directly encode the feature vectors $(\feat_1, \dots, \feat_{i-1})$ into a history embedding $\hemb_i$.
Unlike recurrent models, here the encoding is done independently for each $i$.
The encoding operation can be defined, \eg, using self-attention \cite{zuo2020transformer,zhang2020self}.
It is postulated that such encoders are better at capturing long-range dependencies between events compared to recurrent encoders.
However, more thorough evaluation is needed to validate this claim (see \secref{sec:discussion}).
On the one hand, set aggregation encoders can compute $\hemb_i$ for each event in parallel, unlike recurrent models.
On the other hand, usually the time of this computation scales as $O(N^2)$ with the sequence length $N$, since each $\hemb_i$ depends on all the past events (and the model does not have a Markov property).
This problem can be mitigated by restricting the encoder to only the last $L$ events $(\feat_{i - L}, \dots, \feat_{i - 1})$, thus reducing the time complexity to $O(NL)$.

\subsection{Predicting the Time of the Next Event}
\label{sec:ar-time}
For simplicity, we start by considering the unmarked case and postpone the discussion of marked TPPs until the next section.
An autoregressive TPP models the distribution of the next \emph{arrival} time $t_i$ given the history $\history_{t_i}$.
This is equivalent to considering the distribution of the next \emph{inter-event} time $\tau_i$ given $\history_{t_i}$, which we denote as $\P^*_i(\tau_i)$.\footnote{We again use $*$ to denote conditioning on the history $\history_{t_i}$.}
The distribution $\P^*_i(\tau_i)$ can be represented by any of the following functions:
\begin{enumerate}
    \item probability density function $\pdf_i^*(\tau_i)$
    \item cumulative distribution function $F_i^*(\tau_i) = \int_0^{\tau_i} \pdf_i^*(u)du$
    \item survival function $S_i^*(\tau_i) = 1 - F_i^*(\tau_i)$
    \item hazard function $\hazard_i^*(\tau_i) = \pdf_i^*(\tau_i)/S_i^*(\tau_i)$
    \item cumulative hazard function $\chazard_i^*(\tau_i) = \int_0^{\tau_i}\hazard_i^*(u)du \, .$
\end{enumerate}
In an autoregressive neural TPP, we pick a parametric form for one of the above functions and compute its parameters using the history embedding $\hemb_i$.
For example, the conditional PDF $\pdf_i^*$ might be obtained as
\begin{align}
    \pdf^*_i(\tau_i) = \pdf(\tau_i | \vtheta_i), & & \text{where} & & \vtheta_i = \sigma(\mW \vh_i + \vb).
\end{align}
Here $f(\cdot | \vtheta)$ is some parametric density function over $[0, \infty)$ (\eg, exponential density) and $\mW$, $\vb$ are learnable parameters.
A nonlinear function $\sigma(\cdot)$ can be used to enforce necessary constraints on the parameters, such as non-negativity.

The chosen parametric function should satisfy certain constraints.
First, the inter-event times $\tau_i$ are by definition strictly positive, 
so it should hold that $\smash{\int_{-\infty}^0 f_i^*(u)du = 0}$ or, equivalently, $\chazard_i^*(0) = 0$.
Second, a reasonable default for many applications is to assume that the TPP is \emph{non-terminating}, which means that the next event $t_i$ will arrive eventually with probability one \cite{rasmussen2011temporal}.
This corresponds to the constraint $\int_0^\infty \pdf_i^*(u) du = 1$ or $\lim_{\tau \to \infty}\chazard_i^*(\tau) = \infty$.
The alternative case---a \emph{terminating} point process---means that with a probability $\pi > 0$ no new events will arrive after $t_{i-1}$.
This corresponds to the statement $\int_0^\infty \pdf_i^*(u) du = 1 - \pi$ or, equivalently, $\lim_{\tau \to \infty}\Phi_i^*(\tau) = - \log \pi$.



Specifying one of the functions (1) -- (5) listed above uniquely identifies the conditional distribution $\P^*_i(\tau_i)$, and thus the other four functions in the list.
This, however, does not mean that choosing which function to parametrize is unimportant.
In particular, some choices of the hazard function $\hazard_i^*$ cannot be integrated analytically, which becomes a problem when computing the log-likelihood (as we will see in \secref{sec:learning}). 
In contrast, it is usually trivial to obtain $\hazard_i^*$ from any parametrization of $\chazard_i^*$, since differentiation is easier than integration \cite{omi2019fully}.
More generally, there are three important aspects that one has to keep in mind when specifying the distribution $\P^*_i(\tau_i)$:
\begin{itemize}
    \item \emph{Flexibility:} Does the given parametrization of $\P_i^*(\tau_i)$ allow us to approximate any distribution, \eg, a multimodal one?
    \item \emph{Closed-form likelihood:} Can we compute either the CDF $F_i^*$, SF $S_i^*$ or CHF $\chazard_i^*$ analytically?
    These functions are involved in the log-likelihood computation (\secref{sec:learning}), and therefore should be computed in closed form for efficient model training.
    Approximating these functions with Monte Carlo or numerical quadrature is slower and less accurate.
    \item \emph{Closed-form sampling:} Can we draw samples from $\P_i^*(\tau_i)$ analytically?
    In the best case, this should be done with inversion sampling \cite{rasmussen2011temporal}, which requires analytically inverting either $F_i^*$, $S_i^*$ or $\chazard_i^*$.
    Inverting these functions via numerical root-finding is again slower and less accurate.
    Approaches based on thinning \cite{ogata1981lewis} are also not ideal, since they do not benefit from parallel hardware like GPUs.
    Moreover, closed-form inversion sampling enables the reparametrization trick \cite{mohamed2020monte}, which allows us to train TPPs using sampling-based losses (\secref{sec:learning-alt}).
\end{itemize}
Existing approaches offer different trade-offs between the above criteria.
For example, a simple unimodal distribution offers closed-form sampling and likelihood computation, but lacks flexibility \cite{du2016recurrent}.
One can construct more expressive distributions by parametrizing the cumulative hazard $\chazard_i^*$ either with a mixture of kernels
\cite{okawa2019deep,zhang2020cause,soen2021unipoint} or a neural network \cite{omi2019fully}, but this will prevent closed-form sampling.
Specifying the PDF $\pdf_i^*$ with a mixture distribution \cite{shchur2020intensity} or $\chazard_i^*$ with invertible splines \cite{shchur2020fast} allows to define a flexible model where both sampling and likelihood computation can be done analytically.
Finally, parametrizations that require approximating $\Phi_i^*$ via Monte Carlo integration are less efficient and accurate than all of the above-mentioned approaches \cite{omi2019fully}.

Recently, \cite{soen2021unipoint} have proved that several autoregressive neural TPP models are capable of approximating any other TPP model arbirarily well.
However, in practice such flexible parametrizations listed above might be more difficult to train or more prone to overfitting. Therefore, the choice of the parametrization remains an important modeling decision that depends on the application.

Lastly, we would like to point out that the view of a TPP as an autoregressive model naturally connects to the traditional conditional intensity characterization (\eqref{eq:intensity}).
The conditional intensity $\cintensity(t)$ can be defined by stitching together the hazard functions $\phi_i^*$
\begin{align}
    \label{eq:stitch-hazard}
    \cintensity(t) = \begin{cases}
        \hazard_1^*(t) & \text{if } 0 \le t \le t_1\\
        \hazard_2^*(t - t_1) & \text{if } t_1 < t \le t_2\\
        \vdots\\
        \hazard_{N+1}^*(t - t_N) & \text{if } t_N < t \le T
    \end{cases}
\end{align}
In the TPP literature, the hazard function $\hazard_i^*$ is often called ``intensity,'' even though the two are, technically, different mathematical objects.

\subsection{Modeling the Marks}
\label{sec:ar-marked}
In a \emph{marked} autoregressive TPP, one has to parametrize the conditional distribution $\P^*_i(\tau_i, m_i)$ using the history embedding $\hemb_i$.
We first consider categorical marks, as they are most often used in practice.

\vspace{\parspace}\noindent\textbf{Conditionally independent} models factorize the distribution $\P_i^*(\tau_i, m_i)$ into a product of two independent distributions $\P_i^*(\tau_i)$ and $\P_i^*(m_i)$ 
that are both parametrized using $\hemb_i$ \cite{du2016recurrent,xiao2017modeling}.
The time distribution $\P^*_i(\tau_i)$ can be parametrized using any of the choices described in \secref{sec:ar-time}, such as the hazard function $\phi_i^*(\tau)$. 
The mark distribution $\P^*_i(m_i)$ is a categorical distribution with probability mass function $p_i^*(m_i = k)$.
In this case, the conditional intensity for mark $k$ is computed as
\begin{align}
    \label{eq:int-conditional-ind}
    \cintensity_k(t) = p_i^*(m_i = k) \cdot \phi_i^*(t - t_{i-1}),
\end{align}
where $(i - 1)$ is the index of the most recent event before time $t$.
Note that if we set $K = 1$, we recover the definition of the intensity function from \eqref{eq:stitch-hazard}.

While conditionally independent models require fewer parameters to specify $\P^*_i(\tau_i, m_i)$,
recent works suggest that this simplifying assumption may hurt predictive performance \cite{enguehard2020neural}.
There are two ways to model dependencies between $\tau_i$ and $m_i$ that we consider below.

\vspace{\parspace}
\noindent
\textbf{Time conditioned on marks} \cite{zuo2020transformer,enguehard2020neural}. 
In this case, we must specify a separate distribution $\P_i^*(\tau_i | m_i = k)$ for each mark $k \in \{1, \dots, K\}$.
Suppose that for each $k$ we represent $\P^*_i(\tau_i| m_i = k)$ with a hazard function $\hazard_{ik}^*(\tau)$.
Then the conditional intensity $\cintensity_k(t)$ for mark $k$ is computed simply as
\begin{align}
    \label{eq:time-on-marks}
    \cintensity_k(t) = \phi_{ik}^*(t - t_{i-1}).
\end{align}
It is possible to model the dependencies across marks on a coarser grid in time, which significantly improves the scalability in the number of marks \cite{turkmen2019fastpoint}.

\vspace{\parspace}
\noindent
\textbf{Marks conditioned on time.}
Here, the inter-event time is distributed according to $\P^*_i(\tau_i)$, and for each $\tau$ we need to specify a distribution $\P^*(m_i | \tau_i = \tau)$.
We again assume that the time distribution $\P^*_i(\tau_i)$ is 
described by a hazard function $\phi_i^*(\tau)$, 
and $\P^*_i(m_i | \tau_i = \tau)$ can be parametrized, \eg, using a Gaussian process \cite{bilos2019uncertainty}.
In this case the conditional intensity $\cintensity_k(t)$ is computed as
\begin{align}
    \cintensity_k(t) = p_i^*(m_i = k | \tau_i = t - t_{i-1}) \cdot \phi_i^*(t - t_{i-1}),
\end{align}
where we used notation analogous to \eqref{eq:int-conditional-ind}.
The term $\phi_i^*(t - t_{i-1})$ is often referred to as ``ground intensity.''

\vspace{\parspace}\noindent
\textbf{Other mark types.}
A conditionally independent model can easily handle any type of marks by specifying an appropriate distribution $\P_i^*(m_i)$.
Dependencies between continuous marks and the inter-event time can be incorporated by modeling the joint density $f_i^*(\tau_i, m_i)$ \cite{zhu2020adversarial}.

\section{Continuous-time State Evolution}
\label{sec:continuous-time}
Another line of research has studied neural TPPs that operate completely in continuous time.
Such models define a left-continuous state $\hemb(t)$ at all times $t \in [0, T]$.
The state is initialized to some value $\hemb(0)$.
Then, for each event $i$ the state is updated as
\begin{align}
    \label{eq:cont-time-state}
    \begin{split}
    \vh(t_i) &= \operatorname{Evolve}(\hemb(t_{i-1}), t_{i-1}, t_{i})\\
    \lim_{\varepsilon \to 0}\vh(t_i + \varepsilon) &= \operatorname{Update}(\hemb(t_{i}), \feat_i)
    \end{split}
\end{align}
The $\operatorname{Evolve}(\hemb(t_{i-1}), t_{i-1}, t_i)$ procedure evolves the the state continuously over the time interval $(t_{i-1}, t_i]$ between the events.
Such state evolution can be implemented either via exponential decay \cite{mei2017neural} or, more generally, be governed by an ordinary differential equation \cite{rubanova2019latent,jia2019neural}.
The $\operatorname{Update}(\hemb(t_{i}), \feat_i)$ operation performs an instantaneous update to the state, similarly to the recurrent encoder from last section.

While the above procedure might seem similar to a recurrent model from \secref{sec:ar-encode}, the continuous-time model uses the state $\hemb(t)$ differently from the autoregressive model.
In a continuous-time model, the state $\hemb(t)$ is used to directly define the intensity $\cintensity_k$ for each mark $k$ as
\begin{align}
    \cintensity_k(t) = g_k(\hemb(t)),
\end{align}
where $g_k : \R^H \to \R_{>0}$ is a non-linear function that maps the hidden state $\vh(t)$ to the value of the conditional intensity for mark $k$ at time $t$.
Such function, for example, can be implemented as $g_k(\vh(t)) = \softplus(\vw_k^T \vh(t))$ or a multi-layer perceptron \cite{chen2021neural}.

To summarize, in a continuous-time state model, the state $\vh(t)$ is defined for all $t \in [0, T]$, and the intensity $\cintensity_k(t)$ at time $t$ depends only on the current state $\vh(t)$.
In contrast, in an autogressive model, the discrete-time state $\hemb_i$ is updated only after an event occurs..
Hence, the state $\hemb_i$ defines the entire conditional distribution $\P^*_i(\tau_i, m_i)$, and therefore the intensity $\cintensity_k(t)$ in the interval $(t_i, t_{i+1}]$.

\vspace{\parspace}
\noindent\textbf{Discussion.}
Continuous-time models have several advantages compared to autoregressive ones.
For example, they provide a natural framework for modeling irregularly-sampled time series, which is valuable in medical applications \cite{enguehard2020neural}.
By modeling the unobserved attributes as a function of the state $\hemb(t)$, it is easy to estimate each attribute at any time $t$ \cite{rubanova2019latent}.
These models are also well-suited for modeling spatio-temporal point processes (\ie, with marks in $\gM = \R^D$) \cite{jia2019neural,chen2021neural}.

However, this flexibility comes at a cost: evaluating both the state evolution (\eqref{eq:cont-time-state}) and the model likelihood (\eqref{eq:loglike}) requires numerically approximating intractable integrals.
This makes training in continuous-time models slower than for autoregressive ones.
Sampling similarly requires numerical approximations.

\section{Parameter Estimation}
\label{sec:learning}
\subsection{Maximum Likelihood Estimation}
Negative log-likelihood (NLL) is the default training objective for both neural and classical TPP models.
NLL for a single sequence $X$ with categorical marks is computed as
\begin{align}
    \label{eq:loglike}
    \begin{split}
    -\log p(X) 
    =& -\sum_{i=1}^{N} \sum_{k=1}^{K} \indicator(m_i = k)\log \cintensity_k(t_i)\\
    &+ \sum_{k=1}^{K} \left(\int_{0}^{T} \cintensity_k(u) du\right).
    \end{split}
\end{align}
The log-likelihood can be understood using the following two facts.
First, the quantity $\cintensity_k(t_i)dt$ corresponds to the probability of observing an event of type $k$ in the infinitesimal interval $[t_i, t_i + dt)$ conditioned on the past events $\history_{t_i}$.
Second, we can compute the probability of not observing any events of type $k$ in the rest of the interval $[0, T]$ as $\exp\left(-\int_0^T \cintensity_k(u) du\right)$.
By taking the logarithm, summing these expressions for all events $(t_i, m_i)$ and event types $k$, and finally negating, we obtain \eqref{eq:loglike}.

Computing the NLL for TPPs can sometimes be challenging due to the presence of the integral in the second line of \eqref{eq:loglike}.
One possible solution is to approximate this integral using Monte Carlo integration \cite{mei2017neural} or numerical quadrature \cite{rubanova2019latent,zuo2020transformer}.
However, some autoregressive neural TPP models allow us to compute the NLL analytically, which is more accurate and computationally efficient.
We demonstrate this using the following example.

Suppose we model $\P_i^*(\tau_i, m_i)$ using the ``time conditioned on marks'' approach from \secref{sec:ar-marked}. 
That is, 
we specify the distribution $P_i^*(\tau_i | m_i = k)$ for each mark $k$ with a cumulative hazard function $\chazard_{ik}^*(\tau)$.
By combining \eqref{eq:time-on-marks} and \eqref{eq:loglike}, we can rewrite the expression for the NLL as
\begin{align}
    \label{eq:loglike-haz}
    \begin{split}
    -\log p(X) 
    =& -\sum_{i=1}^{N} \sum_{k=1}^{K} \indicator(m_i = k)\log \phi_{ik}^*(\tau_i)\\
    &+  \sum_{i=1}^{N+1} \sum_{k=1}^{K} \Phi_{ik}^*(\tau_i).
    \end{split}
\end{align}
Assuming that our parametrization allows us to compute $\Phi_{ik}^*(\tau)$ analytically, we are now able to compute the NLL in closed form (without numerical integration).
Remember that the hazard function $\phi_{ik}^*$ can be easily obtained by differentiation as $\phi_{ik}^*(\tau) = \frac{\partial }{\partial \tau}\Phi_{ik}^*(\tau)$.
Finally, note that the NLL can also be expressed in terms of, \eg, the conditional PDFs $\pdf_{ik}^*$ or survival functions $S_{ik}^*$ (\secref{sec:ar-time}).

Evaluating the NLL in \eqref{eq:loglike-haz} can be still computationally expensive when $K$, the number of marks, is extremely large.
Several works propose approximations based on noise-contrastive-estimation that can be used in this regime \cite{guo2018initiator,mei2020noise}.

\vspace{\parspace}
\noindent\textbf{Training.}
Usually, we are given a training set $\dtrain$ of sequences that are assumed to be sampled i.i.d.\ from some unknown data-generating process.
The TPP parameters (\eg, weights of the encoder in \secref{sec:autoregressive}) are learned by minimizing the NLL of the sequences in $\dtrain$.
This is typically done with some variant of (stochastic) gradient descent.
In practice, the NLL loss is often normalized per sequence, \eg, by the interval length $T$ \cite{enguehard2020neural}.
Importantly, this normalization constant cannot depend on $X$, so it would be incorrect to, for example, normalize the NLL by the number of events $N$ in each sequence.
Finally, the i.i.d.\ assumption is not appropriate for all TPP datasets; \cite{boyd2020user} show how to overcome this challenge by learning sequence embeddings.

\subsection{Alternatives to MLE}
\label{sec:learning-alt}
TPPs can be trained using objective functions other than the NLL.
Often, these objectives can be expressed as
\begin{align}
    \label{eq:sampling-loss}
    \E_{X \sim p(X)} \left[f(X)\right].
\end{align}
Such sampling-based losses have been used by several approaches for learning generative models from the training sequences.
These approaches aim to maximize the similarity between the training sequences in $\dtrain$ and sequences $X$ generated by the TPP model $p(X)$ using a scoring function $f(X)$.
Examples include procedures based on Wasserstein distance \cite{xiao2017wasserstein}, adversarial losses \cite{yan2018improving,wu2018adversarial} and inverse reinforcement learning \cite{li2018learning}.

Sometimes, the objective function of the form (\ref{eq:sampling-loss}) 
arises naturally based on the application.
For instance, in reinforcement learning, a TPP $p(X)$ defines a stochastic policy and $f(X)$ is the reward function \cite{upadhyay2018deep}.
When learning with missing data, the missing events $X$ are sampled from the TPP $p(X)$, and $f(X)$ corresponds to the NLL of the observed events \cite{gupta2021learning}.
Finally, in variational inference, the TPP $p(X)$ defines an approximate posterior and $f(X)$ is the evidence lower bound (ELBO) \cite{shchur2020fast,chen2021learning}.

In practice, the gradients of the loss (\eqref{eq:sampling-loss}) w.r.t.\ the model parameters usually cannot be computed analytically and therefore are estimated with Monte Carlo.
Earlier works used the score function estimator \cite{upadhyay2018deep}, but modern approaches rely on the more accurate pathwise gradient estimator (also known as the ``reparametrization trick'') \cite{mohamed2020monte}.
The latter relies on our ability to sample with reparametrization from $P_i^*(\tau_i, m_i)$, which again highlights the importance of the chosen parametrization for the conditional distribution, as described in \secref{sec:ar-time}.

On a related note, sampling-based losses for TPPs (\eqref{eq:sampling-loss}) can be non-differentiable, since $N$, the number of events in a sequence, is a discrete random variable. This problem can be solved by deriving a differentiable relaxation to the loss \cite{shchur2020fast}.


\section{Applications}
\label{sec:applications}

The literature on neural TPPs mostly considers their applications in web-related domains, \eg, for modeling user activity on social media.
Most existing applications of neural TPPs fall into one of two categories:
\begin{itemize}
    \item Prediction tasks, where the goal is to predict the time and / or type of future events;
    \item Structure discovery, where the tasks is to learn dependencies between different event types.
\end{itemize} 
We now discuss these in more detail.

\subsection{Prediction}
Prediction is among the key tasks associated with temporal models.
In case of a TPP, the goal is usually to predict the times and marks of future events given the history of past events.
Such queries can be answered using the conditional distribution $\P_i^*(\tau_i, m_i)$ defined by the neural TPP model.
Nearly all papers mentioned in previous sections feature numerical experiments on such prediction tasks.
Some works combine elements of neural TPP models (\secref{sec:autoregressive}) with other neural network architectures to solve specific real-world prediction tasks related to event data.
We give some examples below.

Recommender systems are a recent application area for TPPs.
Here, the goal is to predict the next most likely purchase event, in terms of both the time of purchase
and the type (\ie, item), given a sequence of customer interactions or purchases in the past.
Neural TPPs are especially well-suited for this task, as they can learn embeddings for large item sets using neural networks, similar to other neural recommendation models.
Moreover, representing the temporal dimension of purchase behavior enables {\em time-sensitive}
recommendations, \eg, used to time promotions.
For example, \cite{kumar2019predicting} address the next item prediction problem by
embedding the event history to a vector, but are concerned only with predicting the next item type (mark).
This approach is extended to a full model of times and events, with a hierarchical RNN model for intra- and
inter-session activity represented on different levels, in \cite{vassoy2019time}. 

Another common application of event sequence prediction is within the {\em human mobility} domain.
Here, events are {\em spatio-temporal}, featuring coordinates in both time and space, potentially along
with other marks. 
Examples include predicting taxi trajectories in a city (\eg, the ECML/PKDD 2015 challenge\footnote{\url{http://www.geolink.pt/ecmlpkdd2015-challenge/}}), or user check-ins on
location-based social media.
\cite{yang2018recurrent} address this task with a full neural TPP, on four different mobility data sets.
This extends the approach in DeepMove \cite{feng2018deepmove}, where the authors similarly use an
RNNs to compute embeddings of timestamped sequences, but limit the predictions to the next location alone.

Other applications include clinical event prediction \cite{enguehard2020neural},
predicting timestamped sequences of interactions of patients with the health system; human activity
prediction for assisted living \cite{shen2018egocentric}, and demand forecasting in sparse time series \cite{turkmen2019intermittent}.

\subsection{Structure Discovery \& Modeling Networks}
In prediction tasks we are interested in the conditional distributions learned by the model.
In contrast, in structure discovery tasks the parameters learned by the model are of interest.

For example, in {\em latent network discovery} applications we observe event activity generated by $K$ users, each represented by a categorical mark.
The goal is to infer an influence matrix $\mA \in \R^{K \times K}$ that encodes the dependencies between different marks \cite{linderman2014discovering}.
Here, the entries of $\mA$ can be interpreted as edge weights in the network.
Historically, this task has been addressed using non-neural models such as the Hawkes process \cite{hawkes1971point}.
The main advantage of network discovery approaches based on neural TPPs \cite{zhang2021learning} is their ability to handle more general interaction types.

Learning \emph{Granger causality} is another task that is closely related to the network discovery problem.
\cite{eichler2017graphical} and \cite{achab2017uncovering} have shown that the influence matrix $\mA$ of a Hawkes process completely captures the notion of Granger causality in multivariate event sequences.
Recently, \cite{zhang2020cause} generalized this understanding to neural TPP models, where they used the method of integrated gradients to estimate dependencies between event types, with applications to information diffusion on the web.

Neural TPPs have also been used to model \emph{information diffusion and network evolution} in social networks \cite{trivedi2019dyrep}.
The DyRep approach by \citeauthor{trivedi2019dyrep} generalizes an earlier non-neural framework, COEVOLVE, by \cite{farajtabar2017coevolve}.
Similarly, Know-Evolve \cite{trivedi2017know} models dynamically evolving knowledge graphs with a neural TPP.
A related method, DeepHawkes, was developed specifically for modeling item popularity in information cascades \cite{cao2017deephawkes}.

\vspace{\parspace}
\noindent\textbf{Other applications.}
Neural TPPs have been featured in works in other research fields and application domains.
For instance, \cite{huang2019recurrent} proposed an RNN-based Poisson process model for speech recognition.
\cite{sharma2018point} developed a latent-variable neural TPP method for modeling the behavior of larval zebrafish.
Performing inference in the model allowed the authors to detect distinct behavioral patterns in zebrafish activity.
Lastly, \cite{upadhyay2018deep} showed how to automatically choose timing for interventions in an interactive environment by combining a TPP model with the framework of reinforcement learning.

\section{Open Challenges}
\label{sec:discussion}
We conclude with a discussion of what, in our opinion, are the main challenges that the field of neural TPPs currently faces.

\subsection{Experimental Setup}
Lack of standardized experimental setups and high-quality benchmark datasets makes a fair comparison of different neural TPP architectures problematic.

Each neural TPP model consists of multiple components, such as the history encoder and the parametrization of the conditional distributions (\twosecrefs{sec:autoregressive}{sec:continuous-time}).
New architectures often change all these components at once,
which makes it hard to pinpoint the source of empirical gains.
Carefully-designed ablation studies are necessary to identify the important design choices and guide the search for better models.

On a related note, the choice of baselines varies greatly across papers.
For example, papers proposing autoregressive models (\secref{sec:autoregressive}) rarely compare to continuous-time state models (\secref{sec:continuous-time}), and vice versa.
Considering a wider range of baselines is necessary to fairly assess the strengths and weaknesses of different families of approaches.

Finally, it is not clear whether the datasets commonly used in TPP literature actually allow us to find models that will perform better on real-world prediction tasks.
In particular, \citeauthor{enguehard2020neural} point out that two popular datasets (MIMIC-II and StackOverflow) can be ``solved'' by a simple history-independent baseline.
Also, common implicit assumptions, such as treating the training sequences as i.i.d.\ (\secref{sec:learning}), might not be appropriate for existing datasets.

To conclude, open-sourcing libraries with reference implementations of various baseline methods and collecting large high-quality benchmark datasets are both critical next steps for neural TPP researchers.
A recently released library with implementations of some autoregressive models by \cite{enguehard2020neural} takes a step in this direction, but, for instance, doesn't include continuous-time state models.

\subsection{Evaluation Metrics}
Many of the metrics commonly used to evaluate TPP models are not well suited for quantifying their predictive performance and have subtle failure modes.

For instance, consider the NLL score (\secref{sec:learning}), one of the most popular metrics for quantifying predictive performance in TPPs.
As mentioned in \secref{sec:ar-marked}, the NLL consists of a continuous component for the time $t_i$ and a discrete component for the mark $m_i$.
Therefore, reporting a single NLL score obscures information regarding the model's performance on predicting marks and times separately.
For example, the NLL score is affected disproportionately by errors in marks as the number of marks increase.
Moreover, the NLL can be ``fooled'' on datasets where the arrival times $t_i$ are measured in discrete units, such as seconds---a flexible model can produce an arbitrarily low NLL by placing narrow density ``spikes'' on these discrete values \cite{uria2013rnade}.

More importantly, the NLL is mostly irrelevant as a measure of error in real-world applications---it yields very little insight into model performance from a domain expert's viewpoint.
However, other metrics, such as accuracy for the mark prediction, and mean absolute error (MAE) or mean squared error (MSE) for inter-event time prediction are even less suited for evaluating neural TPPs.
These metrics measure the quality of a \emph{single} future event prediction, and MAE/MSE only take a \emph{point estimate} of $\tau_i$ into account.
One doesn't need to model the entire distribution over $\tau_i$ (as done by a TPP) to perform well w.r.t.\ such metrics.
If only single-event predictions are of interest, one could instead use a simple baseline that only models $\P_i^*(m_i)$ or produces a point estimate $\tau_i^{\textrm{pred}}$.
This baseline can be trained by directly minimizing absolute or squared error for inter-event times or cross-entropy for marks.
TPPs are, in contrast, probabilistic models trained with the NLL loss; so comparing them to point-estimate baselines using above metrics is unfair.

The main advantage of neural TPPs compared to simple ``point-estimate'' methods is their ability to sample entire trajectories of future events.
Such probabilistic forecasts capture uncertainty in predictions and are able to answer more complex prediction queries 
(\eg, ``How many events of type $k_1$ will happen immediately after an event of type $k_2$?'').
Probabilistic forecasts are universally preferred to point estimates in the neighboring field of time series modeling \cite{gneiting2014probabilistic,alexandrov2020gluonts}.
A variety of metrics for evaluating the quality of such probabilistic forecasts for (regularly-spaced) time series have been proposed \cite{gneiting2008assessing}, but they haven't been generalized to marked continuous-time event sequences.
Developing metrics that are based on entire sampled event sequences can help us unlock the full potential of neural TPPs and allow us to better compare different models.
More generally, we should take advantage of the fact that TPP models learn an entire distribution over trajectories and rethink how these models are applied to prediction tasks.

\subsection{Applications}
While most recent works have focused on developing new architectures for better prediction and structure discovery, other applications of neural TPPs remain largely understudied.

Applications of classical TPPs go beyond the above two tasks.
For example, latent-variable TPP models have been used for event clustering \cite{mavroforakis2017modeling,xu2017dirichlet} and change point detection \cite{altieri2015changepoint}.
Other applications include anomaly detection \cite{li2017detecting}, optimal control \cite{zarezade2017steering,tabibian2019enhancing} and fighting the spread of misinformation online \cite{kim2018leveraging}.

Moreover, most papers on neural TPPs consider datasets originating from the web and related domains, such as recommender systems and knowledge graphs.
Meanwhile, traditional application areas for TPPs, like neuroscience, seismology and finance, have not received as much attention.
Adapting neural TPPs to these traditional domains requires answering exciting research questions.
To name a few, spike trains in neuroscience \cite{aljadeff2016analysis} are characterized by both high numbers of marks (\ie, neurons that are being modeled) and high rates of event occurrence (\ie, firing rates), which requires efficient and scalable models.
Applications in seismology usually require interpretable models \cite{bray2013assessment}, and financial datasets often contain dependencies between various types of assets that are far more complex than self-excitation, commonly encountered in social networks \cite{bacry2015hawkes}.

To summarize, considering new tasks and new application domains for neural TPP models is an important and fruitful direction for future work.

\bibliographystyle{named}
\bibliography{references}

\end{document}

%% file: math_commands.tex
\usepackage{amsmath}
\usepackage{amsfonts}
\usepackage{amssymb}
\usepackage{array}
\usepackage{bbold}
\usepackage{bm}
\usepackage{nicefrac}
\usepackage{xcolor}
\usepackage{xspace}

\newcommand{\history}{\gH}
\newcommand{\feat}{\vy}
\newcommand{\timefeat}{\feat^{\textrm{time}}}
\newcommand{\markfeat}{\feat^{\textrm{mark}}}
\newcommand{\hemb}{\vh}
\newcommand{\hazard}{\phi}
\newcommand{\chazard}{\Phi}
\newcommand{\cintensity}{\lambda^*}
\renewcommand{\P}{P}
\newcommand{\pdf}{f}
\newcommand{\dtrain}{\gD_{\textrm{train}}}

\newcommand{\PreserveBackslash}[1]{\let\temp=\\#1\let\\=\temp}
\newcolumntype{C}[1]{>{\PreserveBackslash\centering}p{#1}}
\newcolumntype{R}[1]{>{\PreserveBackslash\raggedleft}p{#1}}
\newcolumntype{L}[1]{>{\PreserveBackslash\raggedright}p{#1}}

\usepackage{pifont} %
\def\figref#1{Figure~\ref{#1}}

\def\secref#1{Section~\ref{#1}}
\def\twosecrefs#1#2{Sections \ref{#1} and \ref{#2}}

\def\eqref#1{Equation~\ref{#1}}

\newcommand{\indicator}{\mathbb{1}}
\newcommand{\E}{\mathbb{E}}

\newcommand{\R}{\mathbb{R}}

\newcommand{\softplus}{\operatorname{softplus}}

\def\vtheta{{\bm{\theta}}}

\def\vb{{\bm{b}}}

\def\vh{{\bm{h}}}

\def\vw{{\bm{w}}}

\def\vy{{\bm{y}}}

\def\mA{{\bm{A}}}

\def\mW{{\bm{W}}}

\DeclareMathAlphabet{\mathsfit}{\encodingdefault}{\sfdefault}{m}{sl}
\SetMathAlphabet{\mathsfit}{bold}{\encodingdefault}{\sfdefault}{bx}{n}

\def\gD{{\mathcal{D}}}

\def\gH{{\mathcal{H}}}

\def\gM{{\mathcal{M}}}

\newcommand{\ie}{{\em i.e.}}
\newcommand{\eg}{{\em e.g.}}